%% file: main.tex
\documentclass[conference]{IEEEtran}
\IEEEoverridecommandlockouts
\usepackage{cite}
\usepackage{amsmath,amssymb,amsfonts}
\usepackage{algorithmic}
\usepackage{graphicx}
\usepackage{textcomp}
\usepackage{xcolor}

\usepackage{hyperref}
\usepackage{url}

\usepackage{microtype}
\usepackage{graphicx}
\usepackage{booktabs} 
\usepackage{multirow}

\usepackage[full]{complexity}
\usepackage[ruled,vlined]{algorithm2e}
\usepackage{xcolor}   
\usepackage{subfig}
\usepackage{enumitem}

\newcommand{\nop}[1]{}
\DeclareMathOperator*{\argmax}{arg\,max}

\def\BibTeX{{\rm B\kern-.05em{\sc i\kern-.025em b}\kern-.08em
    T\kern-.1667em\lower.7ex\hbox{E}\kern-.125emX}}
\begin{document}

\title{A Generic Graph Sparsification Framework using Deep Reinforcement Learning
}

\author{Ryan Wickman, Xiaofei Zhang, Weizi Li \\
University of Memphis \\
\{rwickman, xiaofei.zhang, wli\}@memphis.edu}

\maketitle

\begin{abstract}
The interconnectedness and interdependence of modern graphs are growing ever more complex, causing enormous resources for processing, storage, communication, and decision-making of these graphs. In this work, we focus on the task \textit{graph sparsification}: an edge-reduced graph of a similar structure to the original graph is produced while various user-defined graph metrics are largely preserved. Existing graph sparsification methods are mostly sampling-based, which introduce high computation complexity in general and lack of flexibility for a different reduction objective. We present SparRL, the \textit{first generic and effective graph sparsification framework} enabled by deep reinforcement learning. SparRL can easily adapt to different reduction goals and promise graph-size-independent complexity. Extensive experiments show that SparRL outperforms all prevailing sparsification methods in producing high-quality sparsified graphs concerning a variety of objectives.
\end{abstract}


\input{sections/intro}
\input{sections/related}

\input{sections/prelim}
\input{sections/method}

\input{sections/experiments}
\input{sections/conclusion}

\bibliographystyle{IEEEtran}
\bibliography{reference,zhang}
\end{document}

%% file: sections/intro.tex
\section{Introduction}
\label{sssec:intro}

Graphs are the natural abstraction of complex correlations found in numerous domains including social media~\cite{borgatti2009network}, communications~\cite{chang2000performance}, transportation and telematics~\cite{Li2017CityFlowRecon,Lin2019ComSense}, and medicine discovery~\cite{barabasi2011network}. Consequently, different types of graphs have been a main topic in many scientific disciplines such as computer science, mathematics, engineering, sociology, and economics. 
The study of the interconnectedness of the graphs can lead to local/global information inferring as well as latent structure discovery, thus benefiting various downstream applications. 
However, the complexity of subject graphs in modern studies tends to ever-increase because of the explosive growth of the Internet and our computing capabilities. These features cause the exploration, analysis, and utilization of a graph very inefficient. 

Thus, the topic \emph{graph sparsification} has emerged in the past two decades~\cite{DBLP:journals/cacm/BatsonSST13, DBLP:journals/fttcs/Teng16},
where the objective is to prune edges from a graph to produce a sparsified graph while preserving user-defined metrics in query evaluation or knowledge inferring. As an example, commonly adopted metrics include the graph spectrum and the effective resistance of edges~\cite{DBLP:journals/siamcomp/SpielmanS11,DBLP:journals/siamcomp/SpielmanT11}. Sparsification techniques developed w.r.t these metrics have been applied to domains such as power grid management~\cite{DBLP:conf/iccad/ZhaoFZ14,DBLP:conf/dac/ZhaoF17}, integrated circuit simulation~\cite{DBLP:journals/tcad/ZhaoHF15}, and influence maximization~\cite{DBLP:conf/sigir/ShenCM17,10.1145/2020408.2020492}. 



Nevertheless, most graph sparsification techniques that exist to date are sampling-based~\cite{DBLP:journals/siamcomp/FungHHP19}. While effective, they all introduce high computation complexity (due to the involvement of many matrix operations) and lack the flexibility to preserve different graph properties in many applications (since the sampling technique needs to be tailored for each application), e.g., approximate graph analysis~\cite{10.1145/3210259.3210269,DBLP:conf/pods/AhnGM12,DBLP:conf/sigmod/SatuluriPR11,10.1145/2806416.2806543}, privacy preserving~\cite{DBLP:conf/asiacrypt/Upadhyay13,DBLP:conf/nips/AroraU19}, and representation learning~\cite{DBLP:conf/icml/CalandrielloKLV18}.    
Thus, a general, flexible graph sparsification technique for various reduction objectives and application domains is highly desired.  

\begin{figure*}%
    \centering
    \small
    \subfloat[\centering Original Karate]{{\includegraphics[height=0.18\textheight,width=0.3\textwidth]{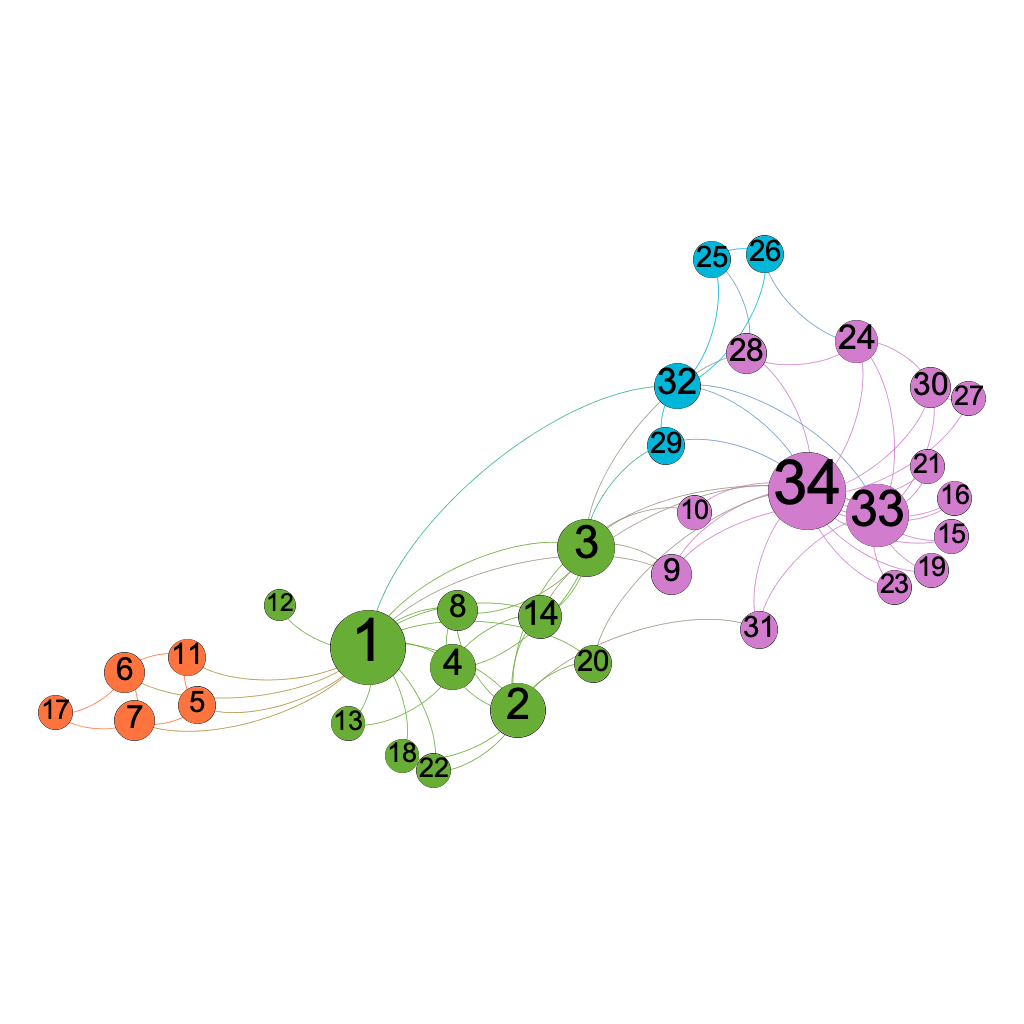} }}%
    \hspace{-2ex}
    \subfloat[\centering Karate (80\% edge kept)]{{\includegraphics[height=0.18\textheight, width=0.3\textwidth]{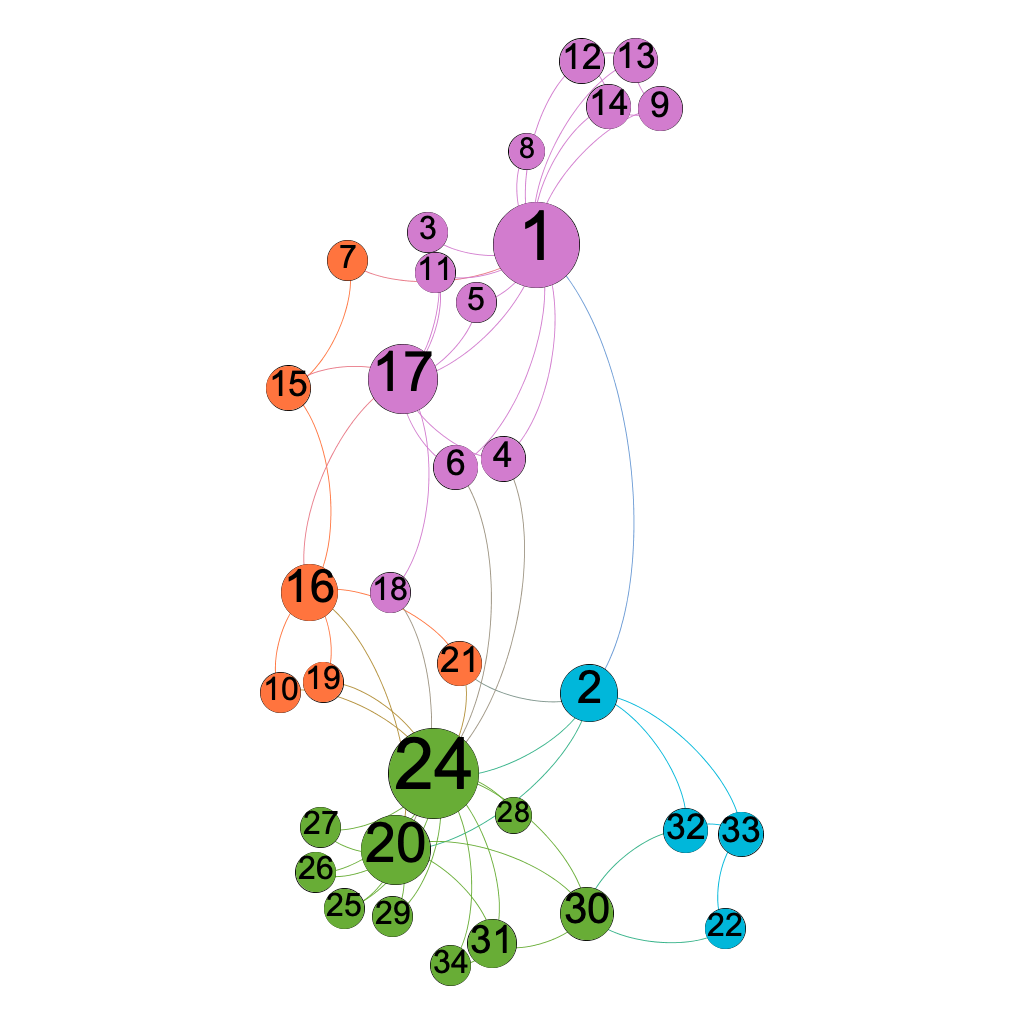} }}\hspace{-4ex}%
    \subfloat[\centering Karate (60\% edge kept)]{{\includegraphics[height=0.18\textheight,width=0.3\textwidth]{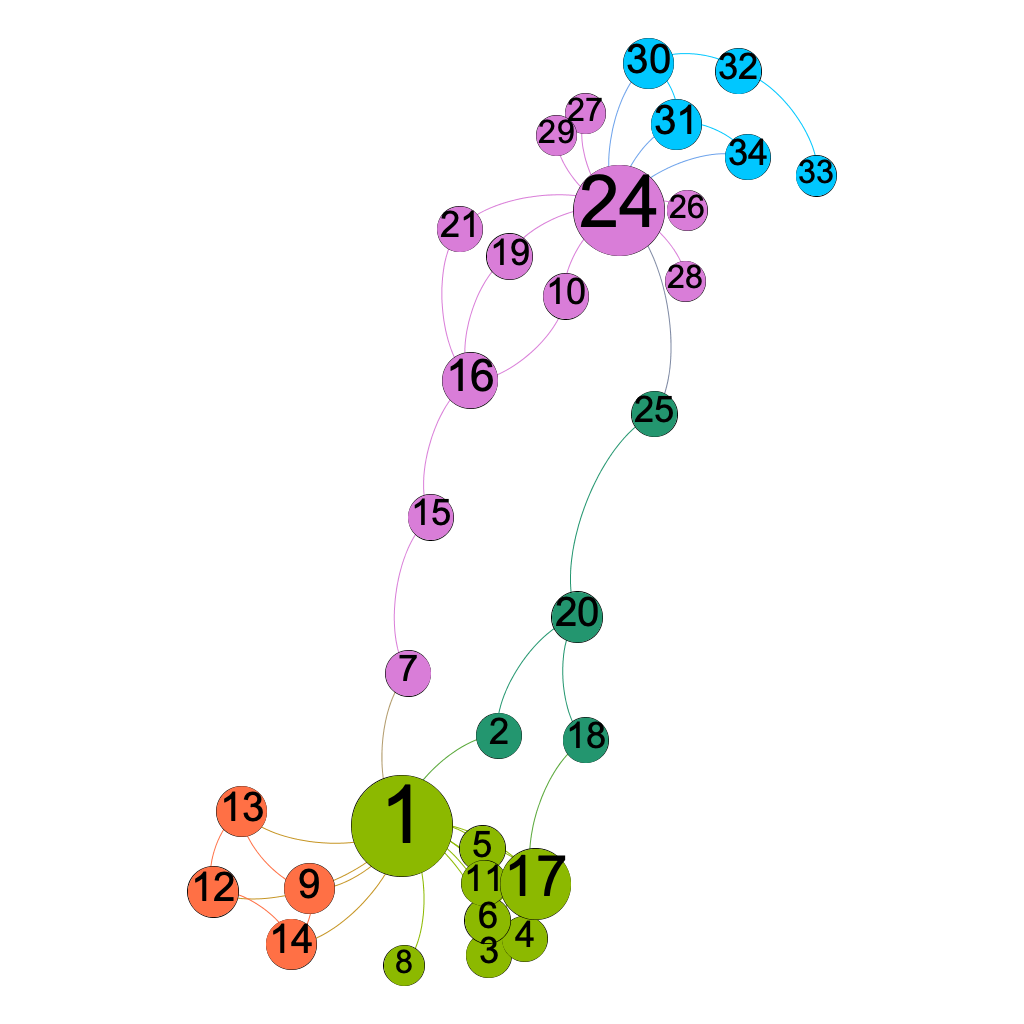} }}%
    \caption{\small Applying SparRL to Zachary's karate club~\cite{karate1977} graph by setting modularity preservation the edge reduction goal. Different colors denote different modularity-based partitions, while a node size scales with its degree. Modularity partition computed and plotted with Gephi~\cite{conf/icwsm/BastianHJ09}.}
    \label{fig:example}%
\end{figure*}

We present SparRL, a general graph sparsification framework empowered by deep Reinforcement Learning (RL) that can be applied to any edge sparsification task with a customized reduction goal. Consider the example shown in Figure~\ref{fig:example}, by setting modularity preservation as the edge reduction objective function, SparRL can prune a user-defined number of edges from the original graph and still preserve the substructure modularity. To improve the learning efficiency and convergence rate of SparRL, we initialize the initial state by randomly sparsifying the graph before each training episode, use Double DQN~\cite{van2016deep} with Prioritized Replay~\cite{schaul2015prioritized}, and employ $\epsilon$-greedy exploration for searching for the optimal pruning strategy. 

We test SparRL using a wide range of graph datasets and metrics including PageRank, community structure, and pairwise shortest-path distance. As a result, SparRL outperforms all baselines on preserving PageRank, community structure, and single-pair shortest path (SPSP) at a variety of edge-kept ratios. 

In summary, the contributions of SparRL are the following: 
\begin{itemize}
    \item A novel reinforcement learning-based general-purpose graph sparsification framework with modular, task-adaptive, and easily modifiable components;
    \item Task flexibility because of the plug-in reward function;
    \item Task scalability as SparRL's time complexity is independent of the size of a graph; and
    \item Simple to use time v.s. performance trade-offs.
\end{itemize}

\noindent The source code of SparRL can be found at \url{https://github.com/rwickman/SparRL-PyTorch}.

%% file: sections/related.tex
\section{Related Work}
\label{sec:relate}
Originated from the cut problem~\cite{DBLP:conf/stoc/BenczurK96}, graph sparsification has drawn extensive research interests~\cite{DBLP:journals/siamcomp/BatsonSS12,DBLP:conf/focs/LeeS15a,DBLP:conf/stoc/Lee017,DBLP:journals/siamcomp/SpielmanS11,DBLP:journals/siamcomp/SpielmanT11}.
A straightforward approach to this problem is to remove edges from a graph with probabilities proportional to the edge weights. However, this approach may fail for a highly structured graph, e.g., a graph with a small cut value where the sampling may exclude the cut edges, thus losing the connectivity of the original graph. An improved approach is to use the k-neighbors sparsifier~\cite{DBLP:conf/aistats/SadhanalaWT16}, which performs local sparsification node-by-node via retaining edges of nodes that have smaller degrees than a predefined threshold $\tau$ and removing edges of nodes (proportionally to their weights) that have larger degrees than $ \tau$. Another method is to remove edges proportionally to their effective resistance, which measures the importance of an edge in preserving the minimum distance between two nodes. This approach will result in only essential edges (on the order of $O(n\polylog n)$) being kept. Although there has been advancement in a fast approximation of effective resistance~\cite{DBLP:conf/stacs/KoutisLP12, DBLP:journals/talg/KoutisLP16}, the graph size remains a dominating factor affecting the computing complexity. 

A related line of research in graph theory, namely graph spanner~\cite{DBLP:journals/siamcomp/PelegU89a}, aims to compute a subgraph that satisfies certain reachability or distance constraints, e.g., $t$-spanner~\cite{DBLP:journals/dam/FeketeK01,DBLP:journals/jcss/DraganFG11}, where the geometric distance of every vertices pair in the subgraph is at most $t$ times of the real distance in the original graph. Therefore, $t$ is also named the stretch parameter, which needs to be specified for a spanner algorithm. However, a spanner has no guarantee on the edge reduction ratio. In contrast, SparRL takes the edge reduction ratio as input and make edge pruning decision via a learned model aiming to best preserve the desired graph structural property. Moreover, techniques developed for the spanner problem are subject to geometric distance preservation, which makes them hard to generalize to a variety of sparsification objectives. 
More details on traditional graph sparsification techniques can be found in survey papers~\cite{DBLP:journals/cacm/BatsonSST13, DBLP:journals/fttcs/Teng16}.


In recent years, learning-based algorithms have gained popularity as they allow for direct optimizing a task's objective and have demonstrated superior performance compared to traditional methods. However, few work exists on learning to sparsify a graph. One related study is GSGAN~\cite{wu2020graph} which approaches graph sparsification using GAN~\cite{goodfellow2014generative}. The goal is to preserve the community structure of a graph by learning to generate a new graph. Despite the effectiveness, GSGAN could introduce edges that do not present in the original graph, thus compromising the graph sparsification objective on many real-world networks where establishing new edges/connections is resource-intensive (e.g., road network). 
Another study, RNet-DQN~\cite{darvariu2020improving}, uses RL to enhance the resilience of graph via adding edges instead of pruning edges. 
GDPNet~\cite{wang2019learning} also uses RL to process graphs, but the goal is representation learning instead of graph sparsification. 

The most relevant study to ours is NeuralSparse~\cite{zheng2020robust}, whose focus is again representation learning and not \textit{generic graph sparsification}. To be specific, NeuralSparse uses two networks to perform a task: the first network produces a sparsified graph via supervised learning; the produced graph then goes through a graph neural network for a downstream classification task. The incurred errors on the classification task are used to tune both networks. While effective, NeuralSparse subjects to two limitations from being a generic graph sparsification tool. First and foremost, the ``quality'' of the sparsified graph cannot be directly assessed but depends on an additional graph neural network. Second, the learning paradigm is limited to classification. Due to these constraints, analytic benchmarks such as node-centrality or shortest-path computing from classical graph sparsification studies are omitted~\cite{zheng2020robust}. In comparison, SparRL outputs a sparsified graph where existing graph analytic benchmarks (and algorithms) can be directly applied. 
To the best of our knowledge, SparRL is the first task-adaptive and effective graph sparsification framework empowered by deep RL.



%% file: sections/prelim.tex
\begin{figure*}
		\centering
		\subfloat[Original Graph]{\includegraphics[scale=0.16]{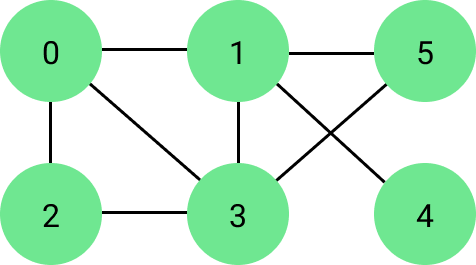}}
		\hspace{5ex}
		\subfloat[SparRL Architecture]{\includegraphics[width=0.58\textwidth]{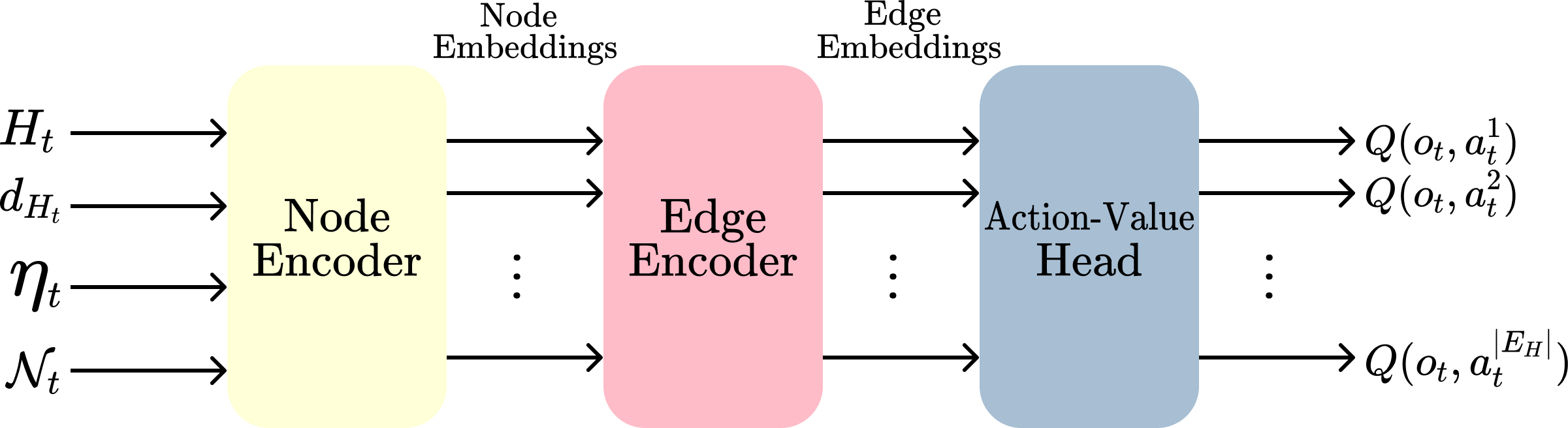}}
		\hspace{5ex}
		\subfloat[Sparsified Graph]{\includegraphics[scale=0.16]{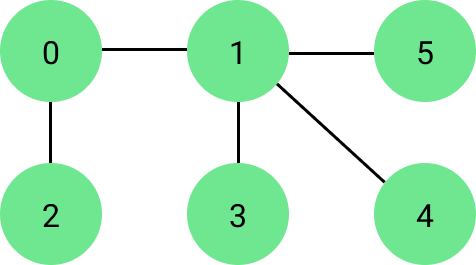}}
		\caption{ \small The SparRL model architecture  consists of the node encoder, edge encoder, and action-value head. The input to the model includes the subgraph $H_t$, the degrees of the nodes $d_{H_t}$, the ratio of edges still in the graph $\eta_t$, and the one-hop neighborhood of the set of nodes in $H_t$. The node encoder uses a GAT~\cite{velivckovic2017graph} on the one-hop neighborhood of each node embedding to create a new node embedding which is then combined with its degrees and $\eta_t$. The edge encoder combines each pair of nodes that represent an edge. The action-value function edge produces the q-value for each edge. }
		\label{fig:sparrlarch}
\end{figure*}

\section{Preliminaries}
\label{sec:prelim}

Given a $T$-step episodic task, at each timestep $t \in [1, T]$, the RL agent uses its policy $\pi_{\theta}(a_t|s_t)$ to choose action $a_t$ based on state $s_t$ from the environment. Then, the environment responds with reward $r_{t}$ and next state $s_{t+1}$. This sequential decision-making process is formulated as a Partially-Observable Markov Decision Process (POMDP) defined by the tuple $(\mathcal{S}, \mathcal{A}, P, R, \Omega, \mathcal{O}, \gamma)$, where $\mathcal{S}$ is the state space, $\mathcal{A}$ is the action space, $P(s'|s,a)$ is the transition probability function, $R$ is the reward function, $\Omega$ is the set of observations, $\mathcal{O}$ is the observation probability function, and $\gamma \in [0,1]$ is the discount factor. 
The objective of the RL agent is to find $\pi_{\theta}$ that can maximize the sum of discounted rewards, or the \emph{return} $R_t = \sum_{i=t}^{T} \gamma^{i-t}r_i$. 

Related to (PO)MDP is the concept of Q-function $Q^\pi: \mathcal{S} \times \mathcal{A} \rightarrow \mathbb{R}$, which describes the total expected reward by taking action $a$ in state $s$ and then following policy $\pi$ thereafter. The element that we want to obtain is the optimal Q-function $Q^*$, from which we can trivially derive the optimal policy $\pi^*(s) = argmax_a Q^*(s,a)$. One way to learn $Q^*$ is through  Q-learning~\cite{watkins1989learning}:

\begin{displaymath}
\begin{split}
    Q_{t+1}(s_t,a_t) \leftarrow & \alpha_t\left[r_{t+1} + \gamma max_{a'} Q_t(s_{t+1},a')-Q_t(s_t,a_t)\right] \\
    & + Q_t(s_t,a_t)  
\end{split}
\end{displaymath}

, where $\alpha_t\in (0,1]$ is the step size. Accompanying the rise of deep learning, many variations of Q-learning have been developed, among which Deep Q-Network (DQN)~\cite{mnih2015human} has gained popularity due to its success in playing Atari games. There exist several major improvements of DQN, including Double DQN~\cite{van2016deep}, Prioritized Replay~\cite{schaul2015prioritized}, and Dueling DQN~\cite{wang2016dueling}. In this work, we choose Double DQN and  Prioritized Replay as a component of SparRL since it has significant improvement over DQN with small code changes.

%% file: sections/method.tex
\section{SparRL Framework}
\label{sec: method}

In this section, we first provide an overview of our approach, then detail its components and design rationale. 

\subsection{Framework Overview}

The aim of this work is to apply a reinforcement learning-based approach to find an edge sparsified graph $G' = (V, E')$, where $V$ represents the set of nodes in the graph and $E'$ represents the set of edges in the sparsified graph, that approximates the original graph $G = (V, E)$, where $E$ is the set of edges in the original graph, measured over some user-defined performance metric.

We treat this as an episodic task, where SparRL sequentially prunes edges from $G$ up until $T$ edges have been removed. Each timestep $t$, an edge-induced subgraph $H_t = (V_{H_t}, E_{H_t})$,  where $V_{H_t}$ represents the set of nodes and $E_{H_t}$ represents the set of edges in the subgraph at timestep $t$, is constructed by sampling a subset of $|E_H|$ edges from the current sparsified graph's edges $E'_t$, thus $V_{H_t} \subseteq V$ and  $E_{H_t} \subseteq E'$. Then, SparRL's action $a_t$ consists of choosing an edge to prune from $E_{H_t}$. This process of sampling and pruning an edge repeats for $T$ timesteps to produce $G'_T$. For ease of notation, we will remove the $t$ subscript whenever we are not referring to an arbitrary timestep in an episode; for example, $|E_H|$ represents the subgraph length for all timesteps, rather than belonging uniquely to a timestep. We describe this process in Algorithm~\ref{alg:framework}. 


\begin{algorithm}
\small
    \caption{SparRL Framework}
    \label{alg:framework}
    \SetKwInOut{Input}{input}\SetKwInOut{Output}{output}
    \Input{$G = (V, E)$, $T$ (the number of edges to prune)}
    \Output{the sparsified graph $G' = (V, E')$ }
\begin{algorithmic}
    \STATE $G' \gets \text{clone }G$
    \FOR{$t=1$ {\bfseries to} $T$}
        \STATE $H_{t} \gets$ Randomly sample a subgraph of $|E_H|$ edges from $G'$
        \STATE $d_{H_t} \gets$ Degrees of nodes in $H_{t}$
        \STATE $\eta_t \gets \frac{|E'_t|}{|E|}$
        \STATE $\mathcal{N}_t \gets$ One-hop neighborhood of nodes in $H_{t}$
        \STATE $q\_values \gets f_{SparRL}(H_{t}, d_{H_t}, \eta_t, \mathcal{N}_t)$
        \STATE $a_t \gets \argmax_{a} q\_values$
        \STATE Prune edge $a_t$ from $G'$
    \ENDFOR
    \RETURN $G'$
\end{algorithmic}
\end{algorithm}



During training, we exploit the simplicity of the environment, by allowing the initial state $s_1$ to be sampled from any state in the state space $\mathcal{S}$. This is implemented as a preprocessing step, where before each episode, we randomly prune $T_p$ edges from $G$ to produce the initial sparsified graph at the first timestep $G'_1$. The number of edges to randomly prune is sampled from a discrete uniform distribution $T_p = \mathcal{U}(0, |E| - T_{max})$, where $T_{max}$ is the maximum number of edges to prune each episode. The upper bound in this case $|E| - T_{max}$ is to prevent pruning more than $|E|$ edges from the graph.

There are a few of benefits of performing this preprocessing step. First, the size of the state space is equivalent to the cardinality of the power set of the edges $|\mathcal{P}(E)|$. So, for any reasonably sized graph, it is trivial to show that it essentially intractable to visit every state. Thus, this step provides sufficient coverage of the space by randomly sampling states to visit. Second, without this step, the requirement for SparRL to reach a particular state with an edge-sparsifed graph $G'$ whose set of edges is $E'$, is to first prune $|E| - |E'|$ edges. Therefore, this requirement is removed, which enables more efficient training. Third, it makes exploration invariant from the behavior of the reinforcement learning policy. This can help prevent early convergence to local optima, as the states that it visits won't be entirely based on the current policy. 

We describe the full training procedure in Algorithm~\ref{alg:training_overview}. This differs from the Algorithm~\ref{alg:framework} as its objective is to train the network, using the preprocessing step just described and other things will go more into detail in later sections, as opposed to getting an edge-sparsified graph at a particular edge-kept ratio.

\begin{algorithm}[t]
\small
    \caption{Training SparRL}
    \label{alg:training_overview}
    \SetKwInOut{Input}{input}\SetKwInOut{Output}{output}
    \Input{$G = (V, E)$, $T_{max}$, $num\_episodes$, $|E_H|$}
\begin{algorithmic}
    \FOR{$i=0$ {\bfseries to} $num\_episodes$}
        \STATE $G' \gets \text{clone }G$
        \STATE Sample $T_p \sim \mathcal{U}(1, |E| - T_{max})$
        \STATE Randomly prune $T_p$ edges from $G'$
        \STATE Sample $T \sim \mathcal{U}(1, T_{max})$
        \FOR{$t=1$ {\bfseries to} $T$}
            \STATE $H_{t} \gets$ Randomly sample a subgraph of $|E_H|$ edges from $G'$
            \STATE $d_{H_t} \gets$ Degrees of nodes in $H_{t}$
            \STATE $\eta_t \gets \frac{|E'_t|}{|E|}$
            \STATE $\mathcal{N}_t \gets$ One-hop neighborhood of nodes in $H_{t}$
            \STATE $q\_values \gets f_{SparRL}(H_{t}, d_{H_t}, \eta_t, \mathcal{N}_t)$
            \STATE $a_t \gets$ Use $\epsilon$-greedy exploration to sample action from $q\_values$ 
            \STATE Prune edge $a_t$ from $G'$
            \STATE $r_t \gets R(G')$
            \STATE Save timestep trajectory $(o_t, o_{t+1}, a_t, r_t)$ in replay buffer
            \STATE Train $f_{SparRL}$ on batch of trajectories from replay buffer
        \ENDFOR
    \ENDFOR
\end{algorithmic}
\end{algorithm}

\subsection{MDP Formulation}

We formalize the task of sparsification as a POMDP that is solved using reinforcement learning. The state space $s_t \in S$ is defined over every possible edge-sparisified graph $G'$, and thus its edges $E'$ can be any element in the power set of the edges in the original graph; $E' \in \mathcal{P}(E)$ and $V$ is assumed to be fixed. The action space $\mathcal{A} \in a_t$ is defined over the set of edges that can be pruned from a given subgraph $H_t$. The reward $r_t \in \mathbb{R}$ is dependent on the properties of the graph that is encouraged to preserve, and thus will be discussed later in the experiments section. The transition probability function $P(s_{t+1} | s_t, a_t)$ is deterministic as the action $a_t$ is fully described by the next state $s_{t+1}$. That is, when the edge referenced by $a_t$ is pruned, the next state is simply $s_{t+1} = G'_{t+1} = (V, E'_t \setminus a_t)$. The observation probability function $\mathcal{O}$ is defined over every possible randomly sampled subgraph $H_t$, and thus assigns a uniform probability to every possible element in this set. The set of observations $o_t \in \Omega$, is defined over what we provide as input into the policy of SparRL. This includes the subgraph $H_t$, the degrees of all the nodes in the subgraph $d_{H_t}$, the ratio of edges left in the graph $\eta_t = |E'_t|/|E|$, and the one-hop neighborhood of the set of nodes in $H_t$, which is $\mathcal{N}_t = \{\left( u, v \right) \in E | u \in V_{H_t} \text{ and } v \in V\}$.

Now, we will provide justification for our input into the policy. The naive choice is to set $o_t = H_t$, which would be a perfectly reasonable choice as input. However, this provides no context on the global state of the graph at the current timestep $G'_t$ beyond $E_{H_t} \subseteq E'_t$. It's trivial to see that many different edge-sparsified graphs could contain this same subgraph. The degrees of the nodes in subgraph $d_{H_t}$ and the one-hop neighborhood $N_t$ are used to inject local context into $o_t$. The ratio of edges left in the graph $\eta_t$ provides some global context. All of these inputs assist the policy by reducing the candidates of possible states of $s_t$, because the policy can identify what set of states corresponds to these properties.

The discovery of the true $s_t$ that corresponds with the current $o_t$ is important, because it helps disambiguate what the true underlying optimal edge sparsification strategy would be in $o_t$. 




\subsection{Policy Learning}
We use Double DQN~\cite{van2016deep} to represent the SparRL sparsification policy that is parameterized by a deep neural network. The policy is trained over a sampled batch of trajectories, sampled using prioritized replay~\cite{schaul2015prioritized}. 



The model architecture, shown in Figure~\ref{fig:sparrlarch}, is composed of the node encoder, edge encoder, and action-value head. The node encoder first looks up the initial node embedding for all the nodes in the graph, which are trained jointly with the model. Then, the node encoder uses a GAT~\cite{velivckovic2017graph} that applies self-attention to the neighborhood of each node to produce a new node embedding. Each node embedding of the subgraph is then separately combined with its degrees, in-degree, and out-degree if the graph is directed, and the ratio of edges left in the graph $\eta_t$. The edge encoder, for every edge, combines the two node embeddings that represent an edge to form an edge embedding. Finally, the action-value head outputs the q-values $Q(o_t, a_t^1)\, \dots \,Q(o_t, a_t^{|E_H|})$, for each edge in the subgraph where $a_t^i$ for $i \in [1, |E_{H}|]$ gives the index of the edge in the subgraph. The GAT in the node encoder uses a single fully-connected layer with 1 unit for computing the attention coefficients, both the main parts of the node encoder and edge encoder consist of two fully-connected layers with 128 units each that are followed by LeakyReLU activation, and the action-value head consists of a single fully-connected layer with one unit.

Thus, the model approximates the Q-value function:
\vspace{-1ex}
\begin{equation}
f_{SparRL}(H_t, d_{H_t}, \eta_t, \mathcal{N}_t) = Q(o_t, a_t^1), \dots, Q(o_t, a_t^{|E_{H}|}). 
\end{equation}

Each edge of the subgraph is independently run through the network, so the subgraph length $|E_H|$ is not constrained by the network. Therefore, any number of edges can be considered to be pruned at each timestep during test time.

\nop{
\begin{figure*}[pt]
		\centering
		\includegraphics[width=0.8\textwidth]{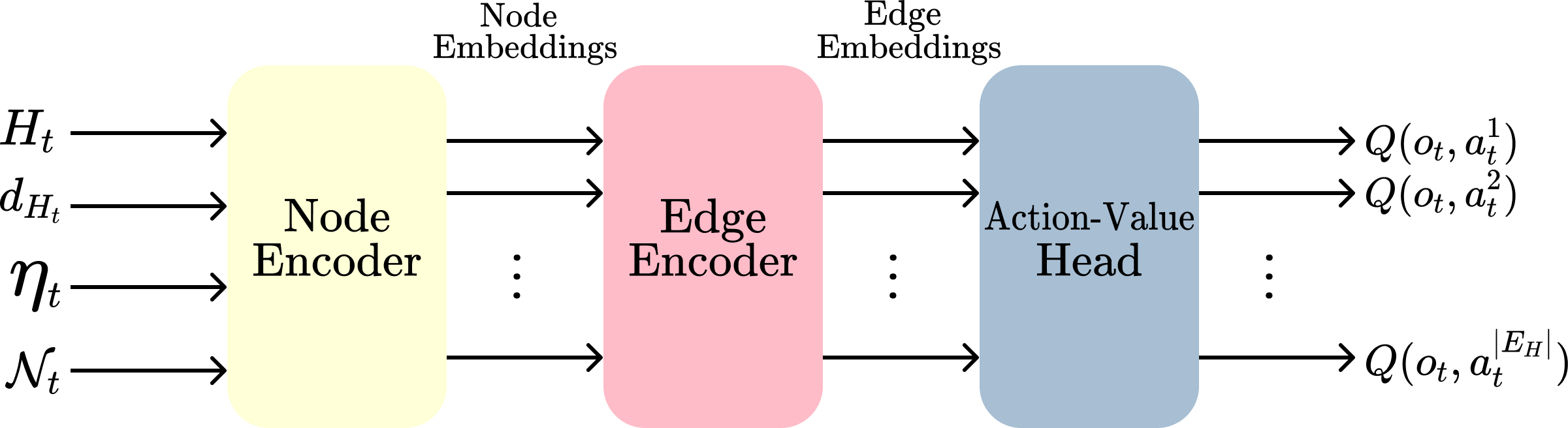}
		\caption{The SparRL model architecture consists of the node encoder, edge encoder, and action-value head. The input to the model includes the subgraph, the degrees of the node, and the ratio of edges still in the graph. The node encoder combine uses a GAT on the 1-hop neighborhood of each node embedding and then combines the output node embedding with its degrees and the ratio of edges left in the graph. The edge encoder will combine each pair of nodes that represent an edge. The action-value function edge will produce the q-value for each edge.} 
		\label{fig:sparrlarch}
\end{figure*}
}

%% file: sections/experiments.tex
\section{Experiments}
\label{sec:exp}

We validate the effectiveness of SparRL using a variety of real-world datasets and test its performance over several metrics. The key observations include: 

\begin{itemize}
    \item SparRL demonstrates superior performance to existing sparsification methods on representative graph metrics over graphs of different scales;
    \item SparRL can outperform the $t$-spanner method for preserving Single-Pair-Shortest-Path (SPSP) distances at the same edge kept ratio; and
    \item SparRL allows for a simple time vs. performance trade-off by modifying $|E_H|$. 
    
\end{itemize}


\begin{table*}[t]
\caption{\small Graph datasets used in the experiments.}
    \centering
   \small
    \tabcolsep=0.1cm
    \begin{tabular}{lcccccc}
        \toprule
        \textbf{Graphs} & \textbf{Twitter} & \textbf{Facebook} & \textbf{YouTube} & \textbf{Amazon } & \textbf{Email} & \textbf{CiteSeer} \\
        \midrule
        $|V|$ & 81,306 & 4,039 & 4,890 & 4,259 & 1,005 & 3,264 \\
        \midrule
        $|E|$ & 1,768,149 & 88,234 & 20,787 & 13,474 & 16,064 & 4,536 \\
        \bottomrule
    \end{tabular}
    \label{tab:datasets-horizon}
\end{table*}
\subsection{Experiment Setup}

\label{exp:setup}
\textbf{Datasets}. We test SparRL using graphs from a variety of domains: Twitter~\cite{fb_cite},  Facebook~\cite{fb_cite}, YouTube~\cite{mislove-2007-socialnetworks} (top-100 communities), Amazon~\cite{yt_cite} (top-500 communities), Email-Eu-Core~\cite{yin2017}, and CiteSeer~\cite{sen2008collective}. Table~\ref{tab:datasets-horizon} summarizes the number of nodes and edges of each graph. 

\nop{
\begin{table}[t]
\caption{Graph datasets used in the experiments.}
\vspace*{-.5em}
    \centering
    \small
    \tabcolsep=0.1cm
    \begin{tabular}{lcc}
        \toprule
        \textbf{Dataset} & Nodes & Edges \\
        \midrule
        Twitter & 81,306 & 1,342,296  \\
        \midrule
        Facebook & 4,039 & 88,234  \\
        \midrule
        YouTube (Top-100 Comm.) & 4,890 & 20,787  \\
        \midrule
        Amazon (Top-500 Comm.) & 4,259 & 13,474   \\
        \midrule
        Email-Eu-Core & 1,005 & 16,064  \\
        \midrule
        CiteSeer & 3,264 & 4,536 \\
        \bottomrule
    \end{tabular}
    \label{tab:datasets}
\end{table}
}


\begin{table*}
\caption{\small Comparison of PageRank preservation via the Spearman's $\rho$ index, where $\eta$ is the edge kept ratio. }
\label{tab:pr_preserve}
\begin{center}
\begin{small}
\scalebox{0.95}{
\begin{tabular}{lcccccccccccc}
\toprule
& \multicolumn{4}{c}{\textbf{Twitter}} & \multicolumn{4}{c}{\textbf{Facebook}} & \multicolumn{4}{c}{\textbf{Amazon (Top-500 Comm.)}}\\
\cmidrule(l){2-5} \cmidrule(l){6-9} \cmidrule(l){10-13}
\textbf{Method} & $\eta$=0.2 & $\eta$=0.4 & $\eta$=0.6 & $\eta$=0.8 & $\eta$=0.2 & $\eta$=0.4 & $\eta$=0.6 & $\eta$=0.8 & $\eta$=0.2 & $\eta$=0.4 & $\eta$=0.6 & $\eta$=0.8\\\cmidrule(l){2-5} \cmidrule(l){6-9} \cmidrule(l){10-13}
SparRL    & \bf{0.846} & \bf{0.944} & \bf{0.984} & \bf{0.995} & \bf{0.942} & \bf{0.982} & \bf{0.996} & \bf{0.998} & \bf{0.779} & \bf{0.907} & \bf{0.944} & \bf{0.988}
\\\cmidrule(l){2-5} \cmidrule(l){6-9} \cmidrule(l){10-13}
LD    & 0.512 & 0.775 & 0.876 & 0.929 & 0.899 & 0.979 & 0.995 & 0.995 & 0.755 & 0.884 & 0.929 & 0.978
\\\cmidrule(l){2-5} \cmidrule(l){6-9} \cmidrule(l){10-13}
RE  & 0.763 & 0.864 & 0.924 & 0.967 & 0.802 & 0.905 & 0.955 & 0.982 & 0.549 & 0.749 & 0.871 & 0.948
\\\cmidrule(l){2-5} \cmidrule(l){6-9} \cmidrule(l){10-13}
EFF    & 0.628 & 0.725 & 0.811 & 0.892 & 0.629 & 0.801 & 0.919 & 0.980 & 0.669 & 0.728 & 0.888 & 0.969    
\\\cmidrule(l){2-5} \cmidrule(l){6-9} \cmidrule(l){10-13}
AD     & 0.520 & 0.690 & 0.837 & 0.940 & 0.408 & 0.519 & 0.637 & 0.782 & 0.230 & 0.341 & 0.553 & 0.769
\\\cmidrule(l){2-5} \cmidrule(l){6-9} \cmidrule(l){10-13}
LS      & 0.771 & 0.826 & 0.857 & 0.892 & 0.648 & 0.830 & 0.924 & 0.960 & 0.589 & 0.640 & 0.763 & 0.859
\\\cmidrule(l){2-5} \cmidrule(l){6-9} \cmidrule(l){10-13}
SB      & 0.581 & 0.689 & 0.761 & 0.811 & 0.379 & 0.582 & 0.681 & 0.740 & 0.247 & 0.348 & 0.397 & 0.399
\\\cmidrule(l){2-5} \cmidrule(l){6-9} \cmidrule(l){10-13}
QSB   & 0.642 & 0.746 & 0.794 & 0.821 & 0.512 & 0.585 & 0.671 & 0.737 & 0.280 & 0.354 & 0.399 & 0.399
\\
\bottomrule
\end{tabular}
}
\end{small}
\end{center}
\vskip -0.1in
\end{table*}

\begin{figure*}[t]
    \centering
    \includegraphics[width=0.7\textwidth]{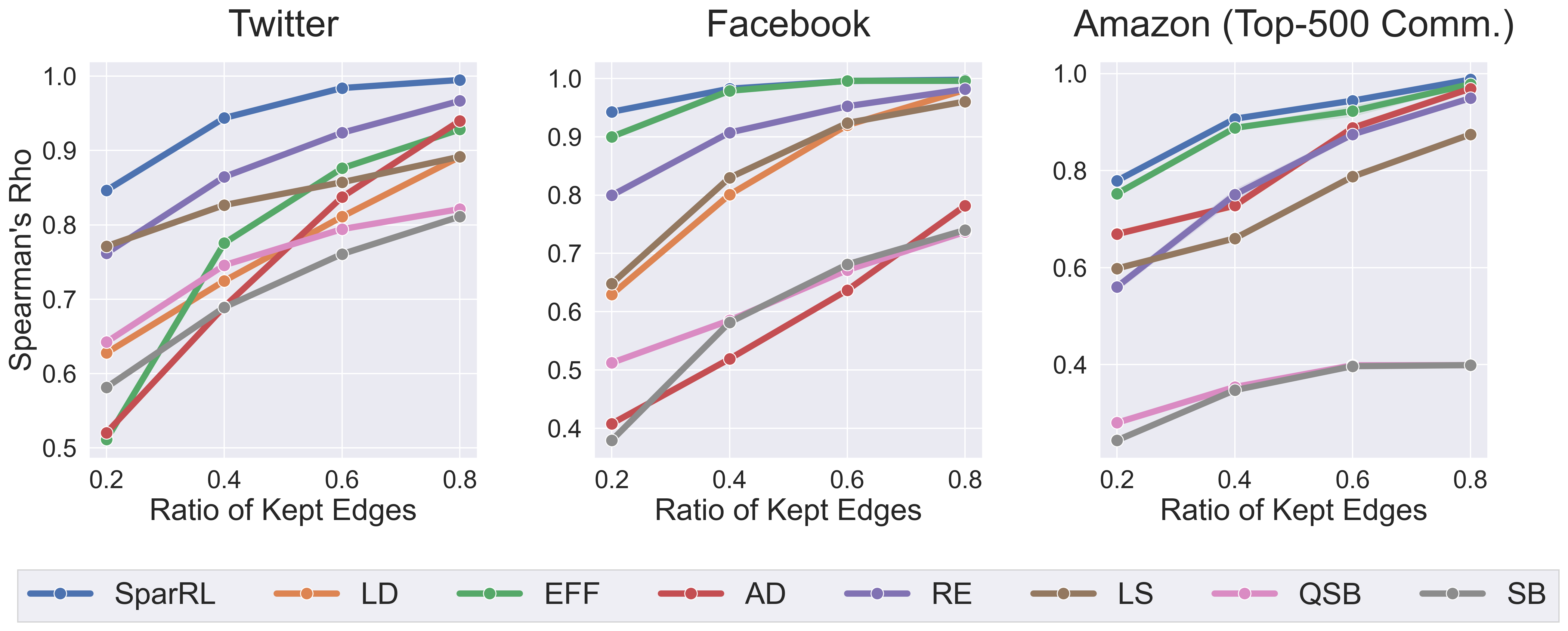}
    \caption{\small PageRank preservation measures over Spearman's $\rho$. SparRL outperforms all other methods on all cases.}\vspace{-2ex}
    \label{fig:pr_preserve}
\end{figure*}

\noindent\textbf{Baseline Methods}. We compare SparRL with a wide range of sparsification methods: 
\begin{itemize}\itemsep0em
\item \textbf{Random Edge (RE)}: RE randomly prunes a given percentage of edges.
\item \textbf{Local Degree (LD)}~\cite{hamann2016structure}: For each node $v \in V$, the edges in the top $\lfloor\text{deg}(v)^{\alpha}\rfloor$ are kept in $G'$, where $\alpha \in [0,1]$.
\item \textbf{Edge Forest Fire (EFF)}~\cite{hamann2016structure}: Based on the Forest Fire node sampling algorithm~\cite{leskovec2006sampling}, a fire is started at a random node and burns approximately $p/(1-p)$ neighbor, where $p$ is the probability threshold of burning a neighbor. Any burnt neighbors are added to a queue to also have a fire started on them. It prunes edges based on the number of times each edge was visited.
\item \textbf{Algebraic Distance (AD)}~\cite{chen2011algebraic}: Based on random walk distance, the algebraic distance $\alpha(u,v)$ between two nodes is low if there is a high probability that a random walk starting from $u$ will reach $v$ using a small number of hops. It uses $1-\alpha(u,v)$ as the edge score so that short-range edges are considered important.
\item \textbf{L-Spar (LS)}~\cite{Satuluri2011}: LS applies the Jaccard similarity function to nodes $u$ and $v$'s adjacency lists to determine the score of edge $(u,v)$. It then ranks edges locally (w.r.t each node) and prunes edges according to their ranks. 
\item \textbf{Simmelian Backbone (SB)}~\cite{nick2013simmelian}: SB measures each edge $(u,v)$'s Simmelianness weight via the shared neighbors of $u$ and $v$. Then, for each $u$, it ranks its neighbors w.r.t the edge weights in descending order. During sparsification, SB will prune each node's lower-ranked edges according to a given edge-prune ratio. 
\item \textbf{Quadrilateral Simmelian Backbone (QSB)}~\cite{nocaj2014untangling}: QSB measures each edge $(u,v)$'s Simmeliannness weight via the shared quadrangles of $u$ and $v$. Then, it follows the same pruning strategy as that of SB.
\end{itemize}

\noindent\textbf{Metrics}. We assess the sparsification methods by examining how well they preserve the topological structure of the original graph w.r.t different metrics, namely PageRank, community structures, and pairwise shortest path distance~\cite{hamann2016structure}. For each metric, we evaluate the performance of a sparsification method by running the method eight times independently and report the average.

\noindent\textbf{SparRL Setup}. Next, we detail our hyperparameter values and settings that are associated with the Double DQN algorithm used by our sparsifier agent. We update the target DQN network by applying soft updates after training on a batch of trajectories each timestep: $\theta_{target}' \gets \varphi\theta_{target} + (1-\varphi)*\theta$, where $\theta_{target}$ are the parameters of the target DQN network and $\theta$ are the parameters of the current DQN network policy. We find that $\varphi=0.005$ corresponds to stable results. We use $\alpha = 0.8$ and $\beta = 0.4$ for Prioritized Replay, $\gamma = 0.99$ for the discounted return, use a batch size of 32, and initially set $\epsilon = 0.99$ for $\epsilon$-greedy exploration and decay it to $0.05$ over the first 10k policy update steps. We keep the learning rate of the model fixed at $0.0002$ during the entire training process.

\begin{table*}
\caption{\small Comparison of community structure preservation over the ARI index, where $\eta$ is the edge kept ratio. }
\label{tab:com_preserve}
\begin{center}
\begin{small}
\scalebox{0.95}{
\begin{tabular}{lcccccccccccc}
\toprule
& \multicolumn{4}{c}{\textbf{YouTube (Top-100 Comm.)}} & \multicolumn{4}{c}{\textbf{Email-Eu-Core}} & \multicolumn{4}{c}{\textbf{Amazon (Top-500 Comm.)}}\\
\cmidrule(l){2-5} \cmidrule(l){6-9} \cmidrule(l){10-13}
\textbf{Method} & $\eta$=0.2 & $\eta$=0.4 & $\eta$=0.6 & $\eta$=0.8 & $\eta$=0.2 & $\eta$=0.4 & $\eta$=0.6 & $\eta$=0.8 & $\eta$=0.2 & $\eta$=0.4 & $\eta$=0.6 & $\eta$=0.8\\\cmidrule(l){2-5} \cmidrule(l){6-9} \cmidrule(l){10-13}
SparRL    & \bf{0.084} & \bf{0.230} & \bf{0.323} & \bf{0.253} & \bf{0.651} & \bf{0.705} & \bf{0.527} & \bf{0.429} & \bf{0.248} & \bf{0.285} & \bf{0.257} & \bf{0.269} 
\\\cmidrule(l){2-5} \cmidrule(l){6-9} \cmidrule(l){10-13}
LD    & 0.052 & 0.082 & 0.150 & 0.145 & 0.278 & 0.30 & 0.278 & 0.207 & 0.240 & 0.238 & 0.236 & 0.236
\\\cmidrule(l){2-5} \cmidrule(l){6-9} \cmidrule(l){10-13}
RE  & 0.048 & 0.144 & 0.157 & 0.152 & 0.226 & 0.318 & 0.348 & 0.312 & 0.141 & 0.224 & 0.236 & 0.235 
\\\cmidrule(l){2-5} \cmidrule(l){6-9} \cmidrule(l){10-13}
EFF    & 0.029 & 0.092 & 0.117 & 0.134 & 0.407 & 0.385 & 0.347 & 0.334 & 0.109 & 0.180 & 0.223 & 0.232    
\\\cmidrule(l){2-5} \cmidrule(l){6-9} \cmidrule(l){10-13}
AD     & 0.035 & 0.068 & 0.111 & 0.143 & 0.263 & 0.272 & 0.284 & 0.333 & 0.105 & 0.169 & 0.210 & 0.234
\\\cmidrule(l){2-5} \cmidrule(l){6-9} \cmidrule(l){10-13}
LS      & 0.016 & 0.071 & 0.150 & 0.147 & 0.446 & 0.390 & 0.375 & 0.319 & 0.209 & 0.250 & 0.249 & 0.245
\\\cmidrule(l){2-5} \cmidrule(l){6-9} \cmidrule(l){10-13}
SB      & 0.039 & 0.127 & 0.161 & 0.166 & 0.274 & 0.326 & 0.365 & 0.358 & 0.225 & 0.225 & 0.235 & 0.249
\\\cmidrule(l){2-5} \cmidrule(l){6-9} \cmidrule(l){10-13}
QSB   & 0.045 & 0.095 & 0.132 & 0.166 & 0.408 & 0.435 & 0.349 & 0.291 & 0.123 & 0.206 & 0.233 & 0.248
\\
\bottomrule
\end{tabular}
}
\end{small}
\end{center}
\vskip -0.1in
\end{table*}

\begin{figure*}[t]
\centering
\includegraphics[width=0.7\textwidth]{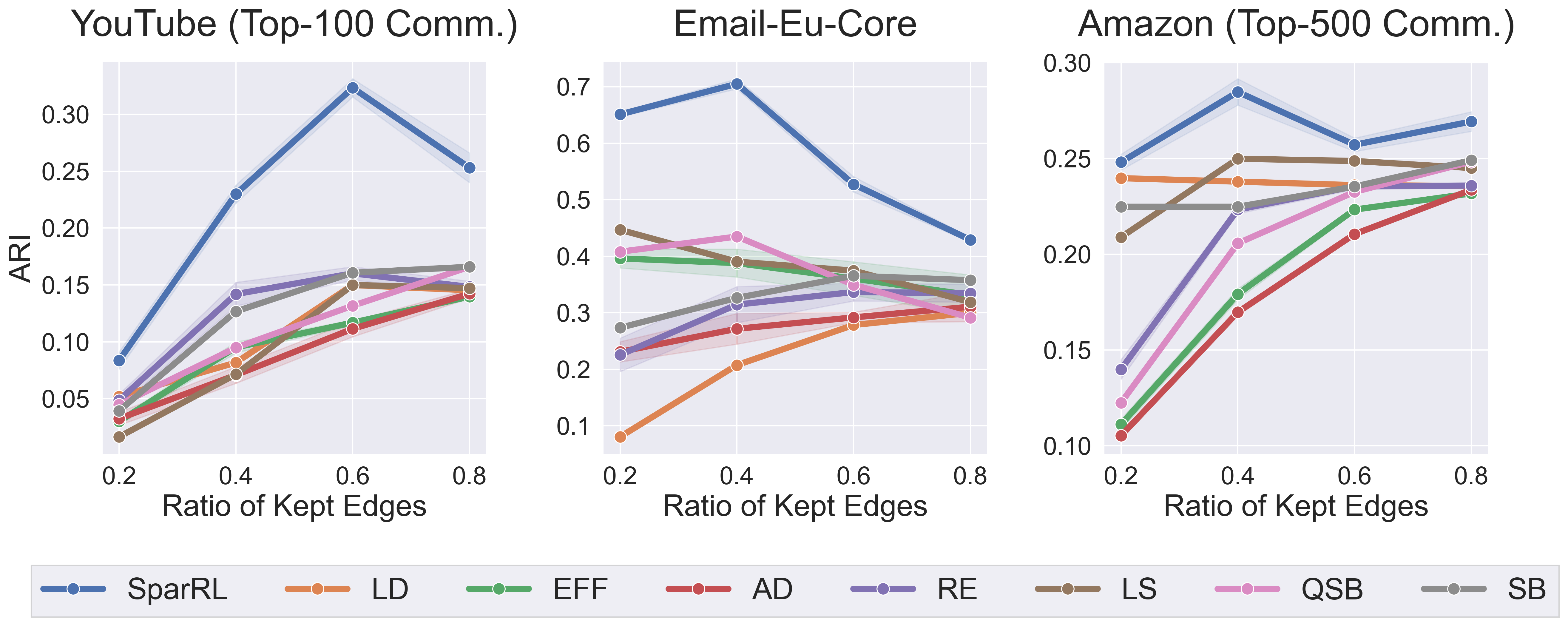}
\caption{\small Community structure preservation measures over ARI score. SparRL outperforms all other methods on all cases.}\vspace{-2ex}
\label{fig:com_preserve}
\end{figure*}

When training the model for all experiments we set $T_{max} = 32$, where we sample $T$ between $[1, T_{max}]$ before each episode and train until negligible improvements are found over pruning $10\%$ of the edges. We set the maximum 1-hop neighborhood of each node to 64, and randomly sample from this set if it is greater to form $\mathcal{N}_t$. Training the SparRL networks typically takes 1h--4h for the smaller graphs and roughly 6h--12h for the larger graphs using an Intel Core i9-11900k CPU and RTX 3090 GPU. However, the run-time complexity is simply $O(|E_H|\times T)$ as we are predicting over a subgraph of length $|E_H|$ for a total of $T$ times. We set $|E_H|$'s default value as 32 and study the impact of $|E_H|$ in Section~\ref{subsec:subgraph_size}.

When evaluating various test datasets, we act greedily w.r.t the learned DQN policy and prune the edge that corresponds to the maximum q-value in the output.

\subsection{Effectiveness of SparRL}

\textbf{PageRank Preservation}. PageRank serves as a critical centrality metric for many ranking-based graph applications. We examine the sparsification methods by comparing the Spearman's $\rho$ rank correlation coefficient~\cite{spearman} between the PageRank score of the original graph and of the sparsified graph at multiple edge-kept ratios, defined as $\eta=|E'|/|E|$.
We define the reward for PageRank as the difference of Spearman's $\rho$ rank correlation coefficients between $G'$ and $G$:

\begin{equation}
r_{pr} = \rho_{G'} - \rho_G,
\end{equation}

where $\rho_{G'}$ is the Spearman's $\rho$ for the sparsified graph $G$ and $\rho_G$ is the Spearman's $\rho$ for the original graph $G$.
We plot the performance comparison of SparRL against all other methods over Twitter, Facebook, and Amazon (Top-500 Comm.) graphs in  Figure~\ref{fig:pr_preserve} and detail all the Spearman's $\rho$ values in Table~\ref{tab:pr_preserve}. It shows that SparRL consistently outperforms other methods at various edge kept ratios ($\eta$), especially when $\eta$ is small.  





\begin{table*}
\caption{\small Comparison of SPSP preservation over the average increase of distance, where $\eta$ is the edge kept ratio. }
\label{tab:spsp_preserve}
\begin{center}
\begin{small}
\scalebox{0.95}{
\begin{tabular}{lcccccccccccc}
\toprule
& \multicolumn{4}{c}{\textbf{Citeseer}} & \multicolumn{4}{c}{\textbf{Email-Eu-Core}} & \multicolumn{4}{c}{\textbf{Amazon (Top-500 Comm.)}}\\
\cmidrule(l){2-5} \cmidrule(l){6-9} \cmidrule(l){10-13}
\textbf{Method} & $\eta$=0.2 & $\eta$=0.4 & $\eta$=0.6 & $\eta$=0.8 & $\eta$=0.2 & $\eta$=0.4 & $\eta$=0.6 & $\eta$=0.8 & $\eta$=0.2 & $\eta$=0.4 & $\eta$=0.6 & $\eta$=0.8\\\cmidrule(l){2-5} \cmidrule(l){6-9} \cmidrule(l){10-13}
SparRL    & \bf{2773} & \bf{2473} & \bf{338} & \bf{15} & \bf{0.326} & \bf{0.133} & \bf{0.07} & \bf{0.021} & \bf{281} &  \bf{106} & \bf{0.518} & \bf{0.24}
\\\cmidrule(l){2-5} \cmidrule(l){6-9} \cmidrule(l){10-13}
LD    & 2896 & 2898 & 2899 & 730 & 0.355 & 0.16 & 0.078 & 0.033 & 299 & 121 & 77 & 9
\\\cmidrule(l){2-5} \cmidrule(l){6-9} \cmidrule(l){10-13}
RE  & 3233 & 2997 & 2047 & 1081 & 309 & 170 & 96 & 41 & 2930 & 1077 & 364 & 117
\\\cmidrule(l){2-5} \cmidrule(l){6-9} \cmidrule(l){10-13}
EFF    & 3064 & 2645 & 1896 & 696 & 173 & 56 & 30 & 18 & 2042 & 1126 & 211 & 56   
\\\cmidrule(l){2-5} \cmidrule(l){6-9} \cmidrule(l){10-13}
AD     & 3252 & 3237 & 3165 & 2995 & 728 & 382 & 240 & 106 & 3803 & 3188 & 2460 & 1396
\\\cmidrule(l){2-5} \cmidrule(l){6-9} \cmidrule(l){10-13}
LS      & 3230 & 3232 & 3204 & 1191 & 107 & 54 & 32 & 31 & 2537 & 813 & 519 & 400
\\\cmidrule(l){2-5} \cmidrule(l){6-9} \cmidrule(l){10-13}
SB        & 3252 & 3192 & 3192 & 3191 & 891 & 531 & 321 & 250 & 2088 & 2084 & 1865 & 1400
\\\cmidrule(l){2-5} \cmidrule(l){6-9} \cmidrule(l){10-13}
QSB   & 3250 & 3192 & 3193 & 3193 & 970 & 465 & 364 & 292 & 3783 & 3022 & 2356 & 1319
\\
\bottomrule
\end{tabular}
}
\end{small}
\end{center}
\vskip -0.1in
\end{table*}

\begin{figure*}[t]
\centering
\includegraphics[width=0.7\textwidth]{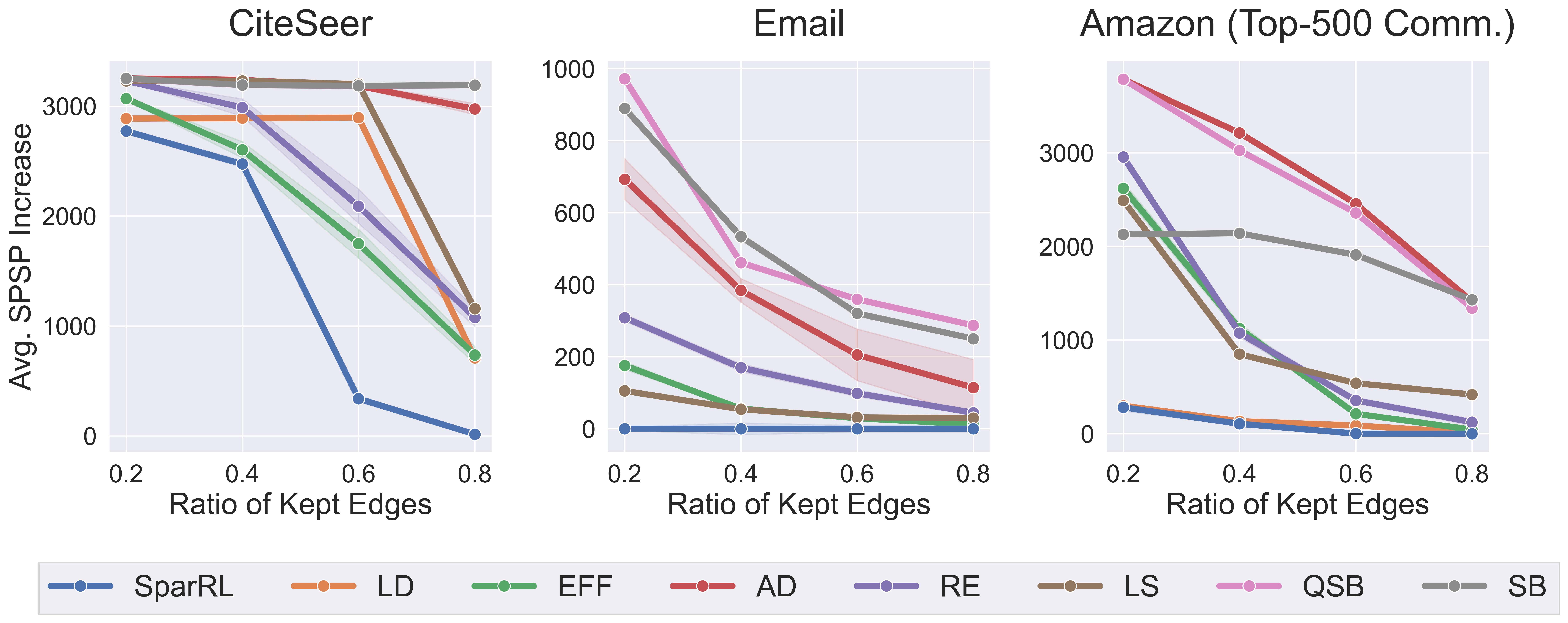}
\caption{\small Shortest Path Distance Preservation measured by the average increase of distance between 8196 randomly selected pairs. In this figure, as SparRL's line is below all the others, it increases the SPSP distance the least and thus outperforms all other methods on all cases.}
\label{fig:spsp_preserve}
\end{figure*}

\textbf{Community Structure Preservation}. We use the Adjusted Rand Index (ARI)~\cite{hubert1985comparing} to measure the effectiveness of SparRL on preserving the community structure of a graph by comparing non-overlapping ground truth communities to those found using the Louvian method~\cite{blondel2008fast} at multiple edge-kept ratios.
We define the reward function as the different between ARI scores for  $G$ and $G'$:

\begin{equation}
r_{com} = ARI(G') - ARI(G) + r_{label},
\end{equation}

where $ARI(G')$ is the Louvian ARI score on $G'$ and $ARI(G)$ is the original Louvian ARI score on $G$.
The other reward, $r_{label}$, is defined as:

\begin{equation}
r_{label} = 
 \begin{cases} 
      1 & l_{a_{t0}} == l_{a_{t1}} \\
     -1 & else 
   \end{cases},
\end{equation}

where $l_{a_{t0}}$ is the label of the source node pruned at timestep t and $l_{a_{t1}}$ is the label of the destination node. We add this auxiliary reward to encourage the agent to not prune an edge if its two nodes belong to the same community.

The results in Figure~\ref{fig:com_preserve} and Table~\ref{tab:com_preserve} show that SparRL consistently outperforms other methods on the YouTube (Top-100 Comm.), Email, and Amazon (Top-500 Comm.) graphs.

\begin{figure*}[t]
    \centering
    \includegraphics[scale=1.0,width=0.7\textwidth ]{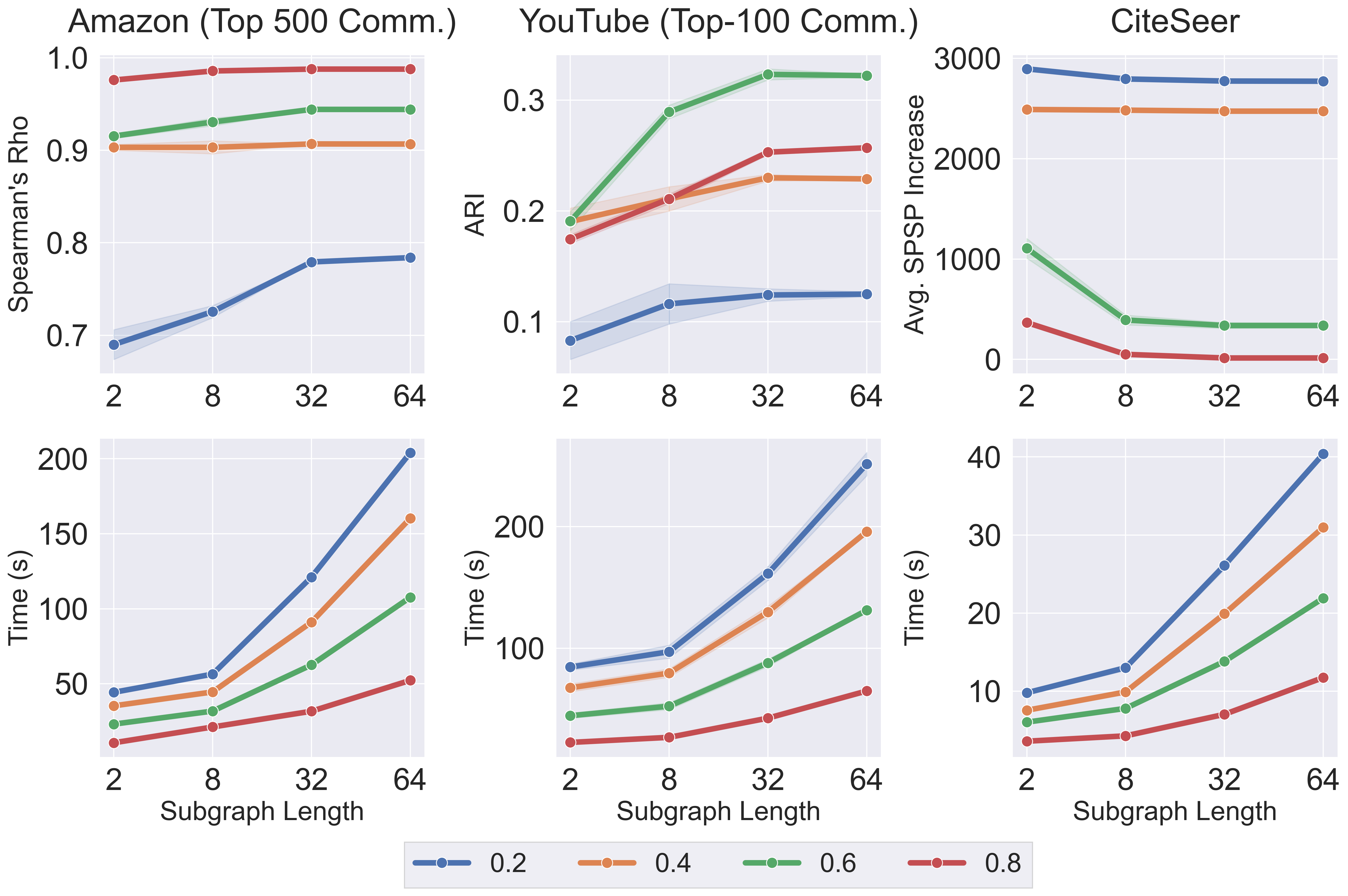}
    \caption{\small Subgraph length $|E_H|$ affect on model performance measured over Spearman's $\rho$, ARI, and Avg. SPSP increase on the Amazon, YouTube, and CiteSeer graphs. It shows that as the subgraph length increases, the performance of the model tends to get better, but at the cost of the increased running time.}\vspace{-2ex}
    \label{fig:subgraph_len_results}
\end{figure*}

\textbf{Shortest Path Distance Preservation}. To test the ability of various sparsification methods in preserving the pairwise shortest path distance, 
We define the reward function for single-pair shortest path (SPSP) as follows:

\begin{equation}
r_{spsp} = \frac{1}{|P|}\sum_{(u,v) \in P } dist(u, v)_{G'} - dist(u, v)_{G},
\label{equ:r_spsp}
\end{equation}

where $P$ is the set of SPSP pairs,  and $dist(u, v)_{G'}$ is the SPSP distance between $u$ and $v$ in the sparsified graph $G'$, and $dist(u, v)_{G}$ is the SPSP distance between node $u$ and node $v$ in the original graph $G$. In the case where $v$ becomes unreachable from $u$ in $G'$, we set this difference equal to $|V|$, as this is greater than the maximum path length by 1. 

During training, the set of SPSPs $P$ is created at each timestep before an edge is pruned. When the model chooses an edge to prune $a_t$, we sample random nodes and compute their shortest paths from the source node and destination node of $a_t$. This design is to leverage the optimal substructure property of shortest paths. That is, the only way a SPSP, for example from $u$ to $v$, will be affected by pruning edge $a_t$ is if it is contained in that path. This is because the SPSP between $u$ and $v$ must be composed of the SPSP from $u$ to the source node in $a_t$ and the SPSP from the destination node in $a_t$ to $v$. Thus, when $a_t$ is pruned, a new SPSP must be generated between $u$ and $v$ that does not contain the edge $a_t$.

During test time, however, we run a maximum of 8196 randomly sampled Single-Pair Shortest Path (SPSP) queries over the graph and keep them fixed for the entire episode. The results are given in Figure~\ref{fig:spsp_preserve} and in Table~\ref{tab:spsp_preserve} which show that SparRL consistently outperforms other methods on the CiteSeer, Email, and Amazon (Top-500 Comm.) graphs.

\begin{table}[t]
\caption{\small SparRL v.s. $t$-spanner for various $t$. ($x\%$: edge kept ratio)}
    \centering
    \small
    \tabcolsep=0.1cm
    \begin{tabular}{lccccc}
        \toprule
        \textbf{Method} & $t$=3   & $t$=4  & $t$=8  & $t$=16  & $t=32$ \\
        & (99.65\%)  & (99.63\%) & (97.82\%) & (93.74\%) & (90.78\%)\\
        \midrule
        t-spanner & 0.0082 & 0.0054 & 0.0405 & 0.1187 & 0.1911\\
        \midrule
        SparRL & \bf{0.0031} & \bf{0.0043} & \bf{0.0350} & \bf{0.0974} & \bf{0.1820} \\
        \bottomrule
    \end{tabular}
    \label{tab:spanner_results}
\end{table}

\textbf{Comparing SparRL and $t$-Spanner}. As $t$-spanner provides a way to sparsify a graph while preserving the geometric distance between a pair of nodes at most $t$ times of the original distance, we conduct an experiment study on comparing the performance of SparRL and a popular spanner algorithm given in~\cite{DBLP:journals/rsa/BaswanaS07}. Due to the fact that spanner algorithms cannot guarantee the number of edges to prune, we run the NetworkX~\cite{SciPyProceedings_11} spanner implementation on various values of stretches and record its edge kept ratio. As it is an approximate algorithm, the spanner algorithm produces a different sparsified graph each run. Therefore, we run the algorithm 16 times for each stretch value and compute the average number of edges and average performance over preserving randomly sampled SPSPs. We then run SparRL on the same average number of edges for each stretch value and display the $r_{spsp}$ (defined in Equation~\ref{equ:r_spsp}) results in Table~\ref{tab:spanner_results} on the CiteSeer network.  These results show that SparRL consistently outperforms the approximate $t$-spanner algorithm over various stretch values. For example, when $t$ equals to 3, $t$-spanner algorithm prunes less than 4\% of edges from the graph. Even with this high edge kept ratio, SparRL delivers a better preservation of the pairwise geometric distance.

\subsection{Impact of Subgraph Size} 
\label{subsec:subgraph_size}
At each timestep, SparRL inputs a subgraph that contains edges randomly sampled from $E'_t$. Due to the flexibility of our model architecture, we can have variable-length subgraphs as input at test time. Thus, in Figures~\ref{fig:subgraph_len_results} we show results on applying the same trained model on varying subgraph lengths and measure their performance over Amazon, YouTube, and CiteSeer w.r.t different metrics, where the shadowed area denotes the standard deviation.  While increasing the subgraph length tends to improve the performance, but the execution time required to prune the graph also increases. Thus, this is a time vs performance trade-off that can be adjusted accordingly based on the user's needs.

%% file: sections/conclusion.tex
\section{Conclusion}
\label{sec:con}
In this work, we propose a general graph sparsification framework based on deep reinforcement learning, namely SparRL. SparRL can overcome the limitations of existing sparsification methods with relatively low computation complexity and the flexibility to adapt to a wide range of sparsification objectives. We evaluate SparRL using various experiments on many real-world datasets and representative graph metrics. The results show that SparRL is effective and generalizable in producing high-quality sparsified graphs. Further analysis of the components of SparRL has validated our design rationale. In the future, we would like to extend SparRL to the dynamic graph setting, and investigate how SparRL could help improve the performance of graph learning tasks, e.g., link prediction, label classification, and other graph-related workloads.


\section*{Acknowledgements} 
This research is partially funded by NSF CCF-2217076 and NSF IIS-2153426. The authors would also like to thank NVIDIA and the University of Memphis for their support.